\documentclass[final]{cvpr}

\usepackage{times}
\usepackage{epsfig}
\usepackage{graphicx}
\usepackage{amsmath}
\usepackage{amssymb}
\usepackage{color}
\usepackage[title]{appendix}

\usepackage{tabulary,multirow,subfig,xspace,xcolor}
\usepackage{amsthm,fixmath,mathtools,nicefrac,mmstyle}

\usepackage[pagebackref=true,breaklinks=true,colorlinks,bookmarks=false]{hyperref}

\begin{document}

%%%%%%%%% TITLE
\title{Seesaw Loss for Long-Tailed Instance Segmentation}

\author{Jiaqi Wang$^{1}$ \quad Wenwei Zhang$^2$ \quad Yuhang Zang$^2$ \quad Yuhang Cao$^{1}$ \quad Jiangmiao Pang$^5$ \quad Tao Gong$^6$ \\
\quad Kai Chen$^{3, 4}$ \quad Ziwei Liu$^2$ \quad Chen Change Loy$^2$ \quad Dahua Lin$^1$ \vspace{5pt} \\
$^1$SenseTime-CUHK Joint Lab, The Chinese University of Hong Kong \\
$^2$S-Lab, Nanyang Technological University \quad $^3$ SenseTime Research \\
$^4$ Shanghai AI Laboratory \quad $^5$Zhejiang University \quad $^6$ University of Science and Technology of China \\
{\tt\small \{wj017,cy020,dhlin\}@ie.cuhk.edu.hk}\hspace{10pt}
{\tt\small \{wenwei001,zang0012,ziwei.liu,ccloy\}@ntu.edu.sg} \hspace{10pt} \\
{\tt\small \{pangjiangmiao,gongtao950513\}@gmail.com} \hspace{10pt}
{\tt\small chenkai@sensetime.com} \hspace{10pt} \\
}

\maketitle

% !TEX root = ../main.tex
\begin{abstract}
	Instance segmentation has witnessed a remarkable progress on class-balanced benchmarks.
	However, they fail to perform as accurately in real-world scenarios, where the category distribution of objects naturally comes with a long tail.
	Instances of head classes dominate a long-tailed dataset and they serve as negative samples of tail categories.
	The overwhelming gradients of negative samples on tail classes lead to a biased learning process for classifiers.
	Consequently, objects of tail categories are more likely to be misclassified as backgrounds or head categories.
	To tackle this problem, we propose Seesaw Loss to dynamically re-balance gradients of positive and negative samples for each category,
	with two complementary factors, \ie, mitigation factor and compensation factor.
	The mitigation factor reduces punishments to tail categories \wrt the ratio of cumulative training instances between different categories.
	Meanwhile, the compensation factor increases the penalty of misclassified instances to avoid false positives of tail categories.
	We conduct extensive experiments on Seesaw Loss with mainstream frameworks and different data sampling strategies.
	With a simple end-to-end training pipeline, Seesaw Loss obtains significant gains over Cross-Entropy Loss,
	and achieves state-of-the-art performance on LVIS dataset without bells and whistles. Code is available at \small{\url{https://github.com/open-mmlab/mmdetection}}.

\end{abstract}

% !TEX root = ../main.tex
\section{Introduction}
Deep learning-based object detection and instance segmentation approaches have achieved immense success
on datasets with relatively balanced category distribution, \eg, COCO dataset~\cite{lin2014coco}.
However, the distribution of categories in the real world is long-tailed~\cite{OLTR}. There are
a few head classes containing abundant instances, while most other classes comprise relatively few instances.

\begin{figure}
    \centering
    \includegraphics[width=\linewidth]{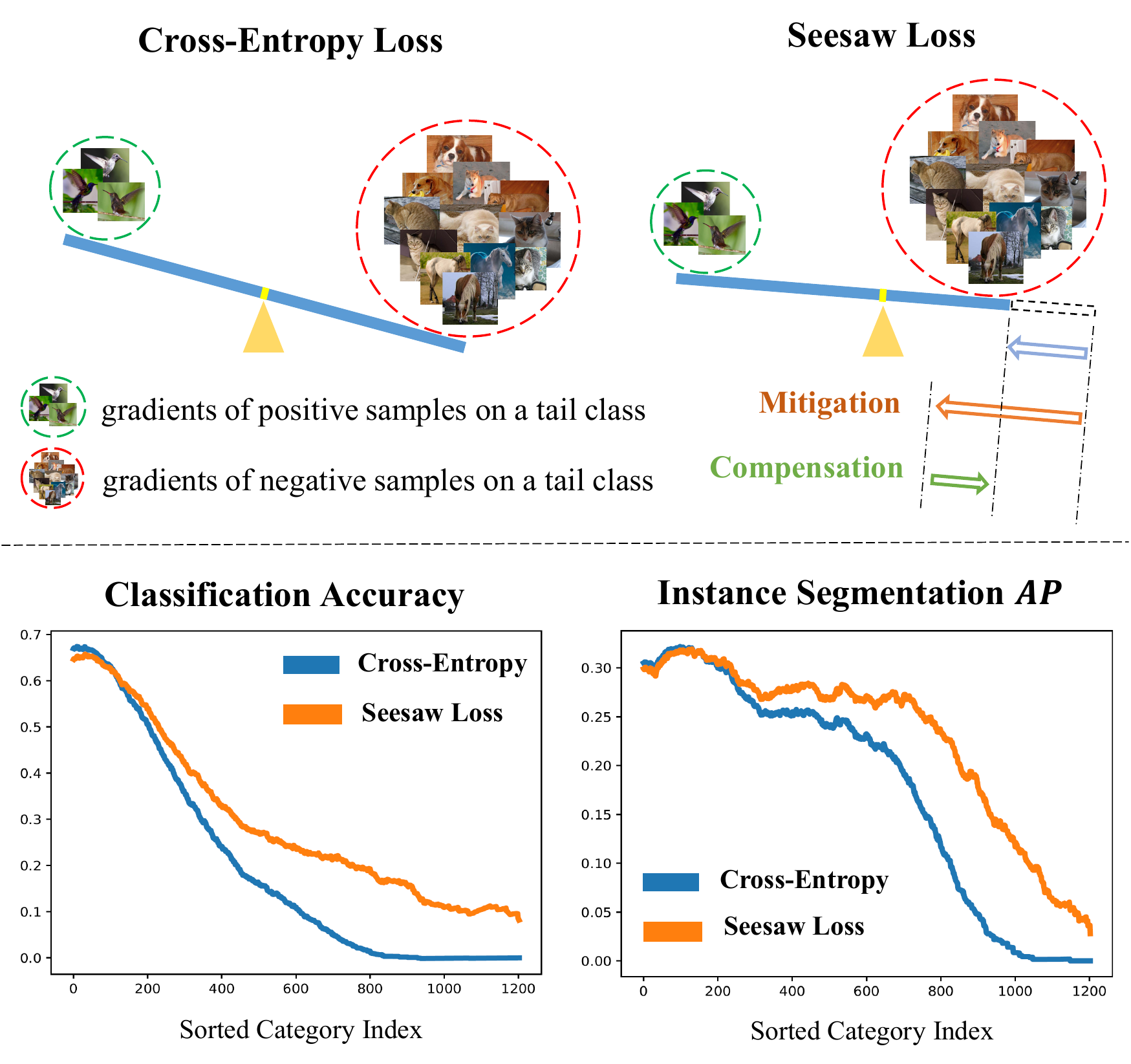}
    \vspace{-18pt}
    \caption{\small{Seesaw Loss dynamically re-balances the gradients of positive and negative samples on a tail class with two complementary factors.
            It mitigates the overwhelming punishments on the tail class as well as compensates them to reduce the risk of inducing false positives.
            In Mask R-CNN~\cite{He_2017}, Seesaw Loss achieves remarkable higher classification accuracy of tail classes
            than Cross-Entropy Loss on LVIS~\cite{gupta2019lvis} dataset.
            As a result, instance segmentation $AP$ on tail classes is significantly improved, leading to better overall performance.
        }
    }
    \vspace{-18pt}
    \label{fig:teaser1}
\end{figure}

On long-tailed datasets, existing instance segmentation frameworks~\cite{cai2019cascadercnn,chen2019hybrid, He_2017}
fail to perform as accurately as on the datasets with balanced category distribution, exhibiting unsatisfactory performance on tail classes.
Figure~\ref{fig:teaser1} shows the classification accuracy and instance segmentation performance of Mask R-CNN~\cite{He_2017} on LVIS~\cite{gupta2019lvis} dataset.
The classifier in Mask R-CNN trained by Cross-Entropy Loss tends to misclassify tail categories as backgrounds or other confusing head classes,
which leads to extremely low accuracy on tail classes.

The primary reason for this undesired phenomenon is that the instances from head classes are predominant in a long-tailed dataset.
These instances contribute an overwhelmingly large quantity of negative samples for tail classes.
Thus, the gradients of positive and negative samples on a tail class
are heavily imbalanced, leading to a biased learning process for the classifier.
One can imagine that gradients of positive and negative samples resemble two objects positioned on each end of a seesaw (see Fig.~\ref{fig:teaser1}).
To balance them, a viable solution is to shorten the arm of the heavier end in the seesaw,
which is equivalent to scaling down the overwhelming gradients of negative samples on the tail class by a factor.
Nevertheless, blindly reducing the gradients of negative samples increases the risk of inducing false positives of tail classes,
since samples of other classes are less punished when they are misclassified as tail classes.
Thus, a specialized mechanism is needed to compensate for the excessively reduced penalties on tail classes.

In this work, we propose \textbf{Seesaw Loss} that dynamically re-balances positive and negative gradients for each category with two complementary factors, \ie, mitigation factor and compensation factor.
According to the ratio between categories' cumulative sample numbers during training,
the mitigation factor reduces the penalty to relatively rare classes.
When a false positive sample of one category is observed, the compensation factor will increase the penalty to that category.
The synergy of the two above factors enables Seesaw Loss to mitigate the overwhelming punishments to tail classes as well as compensate for the risk of
misclassification caused by diminished penalties.

Seesaw Loss has three appealing properties.
\textbf{1)} Seesaw Loss is \textbf{dynamic}.
It explores the ratios of cumulative training sample numbers between different categories
and instance-wise misclassification during training.
This differs significantly to previous solutions that rely either on static group split~\cite{groupsoftmax} or
loss reweighting with constant values~\cite{cao_2019_LDAM, Cui_2019, EQL}.
\textbf{2)} Seesaw Loss is \textbf{self-calibrated}.
The mitigation and the compensation factor synergize to relieve the
overwhelming punishments on tail classes as well as avoid increasing false positives of tail categories.
On the contrary, previous methods blindly reduce punishments on tail classes~\cite{EQL}
or decrease the loss weights of head categories~\cite{Cui_2019}.
\textbf{3)} Seesaw Loss is \textbf{distribution-agnostic}.
It does not rely on pre-computed datasets' distribution ~\cite{cao_2019_LDAM,Cui_2019,OLTR,EQL},
and it can operate well with any data sampler~\cite{gupta2019lvis,Hu_2020_CVPR}.
By accumulating the number of samples in each class, Seesaw Loss gradually approximates the real data distribution during training to achieve more accurate balancing.

Through extensive experiments, we show consistent improvements of Seesaw Loss in different instance segmentation frameworks and data samplers.
On the challenging LVIS~\cite{gupta2019lvis} dataset, Seesaw Loss
achieves significant improvements of
6.0\% $\text{AP}$ and 2.1\% $\text{AP}$ upon Mask R-CNN~\cite{He_2017} with random sampler and repeat factor sampler~\cite{gupta2019lvis}, respectively.
Even if switching to the stronger Cascade Mask R-CNN~\cite{cai2019cascadercnn},
we still observe an impressive improvement of 6.4\% $\text{AP}$ and 2.3\% $\text{AP}$
with random sampler and repeat factor sampler.
To show the versatility of Seesaw Loss, we integrate it into the long-tailed image classification task.
Seesaw Loss significantly improves the classification accuracy by 6\% on ImageNet-LT~\cite{OLTR} dataset.
Besides, we also explore the necessity of the decoupling training pipeline~\cite{Kang2020Decoupling, groupsoftmax} in Seesaw Loss.
Experimental results demonstrate that Seesaw Loss
provides a simpler and more effective solution to long-tailed instance segmentation without relying on complex training pipelines.

% !TEX root = ../main.tex
\section{Related Work}
\noindent\textbf{Object Detection.}
Recent years have witnessed a remarkable improvement in object detection~\cite{nas_fpn, He_2019_ICCV, SABL,Wang_2019_ICCV,wang2020carafe}.
A leading paradigm in this area is the two-stage pipeline~\cite{girshick2015fast, ren2015faster}, where
the first stage generates a set of region proposals, and then the second stage classifies and refines the proposals.
Unlike the two-stage approaches, the single-stage pipeline~\cite{lin2017_focal, liu2016_ssd, Redmon_2016, Redmon_2017} directly predicts bounding boxes.
Classical single-stage approaches ~\cite{lin2017_focal, liu2016_ssd} require densely populated anchors as a prior,
while anchor-free methods~\cite{kong2019foveabox, Law2018_CornerNet, tian2019fcos, wang2019nasfcos}
manage to achieve similar or better performance without such prior.
There are also attempts to apply cascade architecture~\cite{cai2019cascadercnn, Gidaris_2016, jiang2018acquisition, najibi2016g, wang2019region}
to refine the bounding boxes' predictions progressively.

\noindent\textbf{Instance Segmentation.}
Instance segmentation is becoming popular in tandem with a surge in the interest in object detection.
Early methods perform segmentation before object recognition~\cite{DeepMask, RefineSegments}.
Via adding a mask prediction branch in the Faster R-CNN~\cite{ren2015faster} architecture,
Mask R-CNN~\cite{He_2017} bridges the gap between object detection and instance segmentation.
The idea is also adopted by ~\cite{cai2019cascadercnn, chen2019hybrid} in their cascading frameworks.
More recent works~\cite{TensorMask, SOLO} introduce an even shorter pipeline by skipping the detection process and directly predicting mask for each instance.
Seesaw Loss can easily cooperates with object detection and instance segmentation frameworks for the long-tailed datasets.

\noindent\textbf{Long-Tailed Recognition.}
Long-tailed recognition tasks~\cite{OLTR, gupta2019lvis, Learn_imbalance, wu2020distribution} receive growing attention
recently as the problems are closer to real-world applications.
One representative solution to the problem is loss re-weighting~\cite{cao_2019_LDAM, Cui_2019,  Huang_imbalance_2020}.
Loss re-weighting methods adopt different re-weighting strategies~\cite{cao_2019_LDAM, Cui_2019, Huang_imbalance_2020, lin2017_focal, EQL}
to adjust the loss of different classes based on each class's statistics~\cite{cao_2019_LDAM, Cui_2019}.
Other common approches~\cite{Learn_imbalance, Hu_2020_CVPR, Mahajan_explore} re-balance the distribution of the instance numbers in each class,
\eg, repeat factor sampling \cite{gupta2019lvis} and class-balanced sampling~\cite{Mahajan_explore}, both are based on the sample numbers of classes.
Different sampling strategies can be adopted at different training stages to formulate a multi-stage training procedure~\cite{Hu_2020_CVPR, Kang2020Decoupling}.
A recent work~\cite{Kang2020Decoupling} proposes a decoupling training pipeline. which first trains a good representation network with natural sampling and then finetunes
the classifier with class-balanced sampling.
There are also attempts to modify the classifier to improve the performance on tail classes,
\eg, using different classifiers for different groups of classes~\cite{groupsoftmax},
or use two classifiers trained with different data samplers~\cite{wang2020devil}.

% !TEX root = ../main.tex
\section{Methodology} \label{sec:method}
The classifier trained by the widely applied Cross-Entropy (CE) Loss (Sec.~\ref{sec:celoss}) is highly biased on long-tailed datasets,
resulting in much lower accuracy of tail classes than head classes.
The major reason is that gradients brought by positive samples are overwhelmed by gradients from negative samples on tail classes.
Therefore, we propose Seesaw Loss to mitigate the overwhelming gradients of negative samples on tail classes as well as
compensate the gradients of misclassified samples to avoid false positives (Sec.~\ref{sec:seesawloss}).
We also explore some practical component designs to adopt Seesaw Loss in instance segmentation (Sec.~\ref{sec:modeldesign}).

\subsection{Cross-Entropy Loss}\label{sec:celoss}
We first revisit the most widely adopted Cross-Entropy (CE) Loss in existing frameworks~\cite{chen2019hybrid, He_2017}.
The formulation of CE Loss can be written as
\begin{equation} \label{eq:celoss}
	\begin{aligned}
		L_{ce}(\vz)=-\sum_{i=1}^{C} y_{i} \log (\sigma_{i}), \quad \text{ with } \sigma_{i}=\frac{e^{z_{i}}}{\sum_{j=1}^{C}e^{z_{j}}},
	\end{aligned}
\end{equation}
where $\vz=[z_1, z_2, \dots, z_C]$ and $\vsigma = [\sigma_1, \sigma_2, \dots, \sigma_C]$ are the predicted logits and probabilities of the classifier, respectively.
And $y_i \in \{0,1\},
	1\leq i \leq C$ is the one-hot ground truth label.
Given a training sample of class $i$, the gradients on $z_i$ and $z_j$ are given by
\begin{equation}
	\frac{\partial L_{ce}(\vz)} {\partial z_i} = \sigma_i - 1,
\end{equation}
\begin{equation}\label{eq:ceng}
	\frac{\partial L_{ce}(\vz)} {\partial z_j} =  \sigma_j,
\end{equation}
It shows that samples of class $i$ punish the classifier of class $j$ \wrt $\sigma_j$.
In the case that the instance number of class $i$ is enormously greater than that of class $j$,
the classifier of class $j$ will receive penalties in most samples and attains few positive signals during training.
Thus the predicted probabilities of class $j$ will be heavily suppressed, which results in a low classification accuracy of tail classes, as shown in Figure~\ref{fig:teaser1}.

\subsection{Seesaw Loss}\label{sec:seesawloss}
To alleviate the above mentioned problem, one feasible solution is to decrease the gradients of negative samples
in Eq.~\ref{eq:ceng} imposed by head classes on a tail class.
Therefore, we propose Seesaw Loss as
\begin{equation} \label{eql:seesawloss}
	\begin{aligned}
		L_{seesaw}(\vz)=-\sum_{i=1}^{C} y_{i} \log (\widehat{\sigma}_{i}), \\
		\text{  with  } \widehat{\sigma}_{i}=\frac{e^{z_{i}}}{\sum_{j\neq i}^{C}\cS_{ij}e^{z_{j}}+e^{z_{i}}}.
	\end{aligned}
\end{equation}
Then the gradient on $z_j$ of negative class $j$ in Eqn~\ref{eq:ceng} becomes
\begin{equation}
	\frac{\partial L_{seesaw}(\vz)} {\partial z_j} = \cS_{ij} \frac{e^{z_j}} {e^{z_i}} \widehat{\sigma}_i.
\end{equation}
Here $\cS_{ij}$ works as a tunable balancing factor between different classes.
By a careful design of $\cS_{ij}$,
Seesaw loss adjusts the punishments on class $j$ from positive samples of class $i$.
Seesaw loss determines $\cS_{ij}$ by a mitigation factor and a compensation factor, as
\begin{equation}
	\cS_{i j} = \cM_{i j} \cdot \cC_{ij}.
\end{equation}
The mitigation factor $\cM_{i j}$ decreases the penalty on tail class $j$ according to a ratio of instance numbers between tail class $j$ and head class $i$.
The compensation factor $\cC_{ij}$ increases the penalty on class $j$ whenever an instance of class $i$ is misclassified to class $j$.

\begin{figure}[t]
	\begin{center}
		\includegraphics[width=\linewidth]{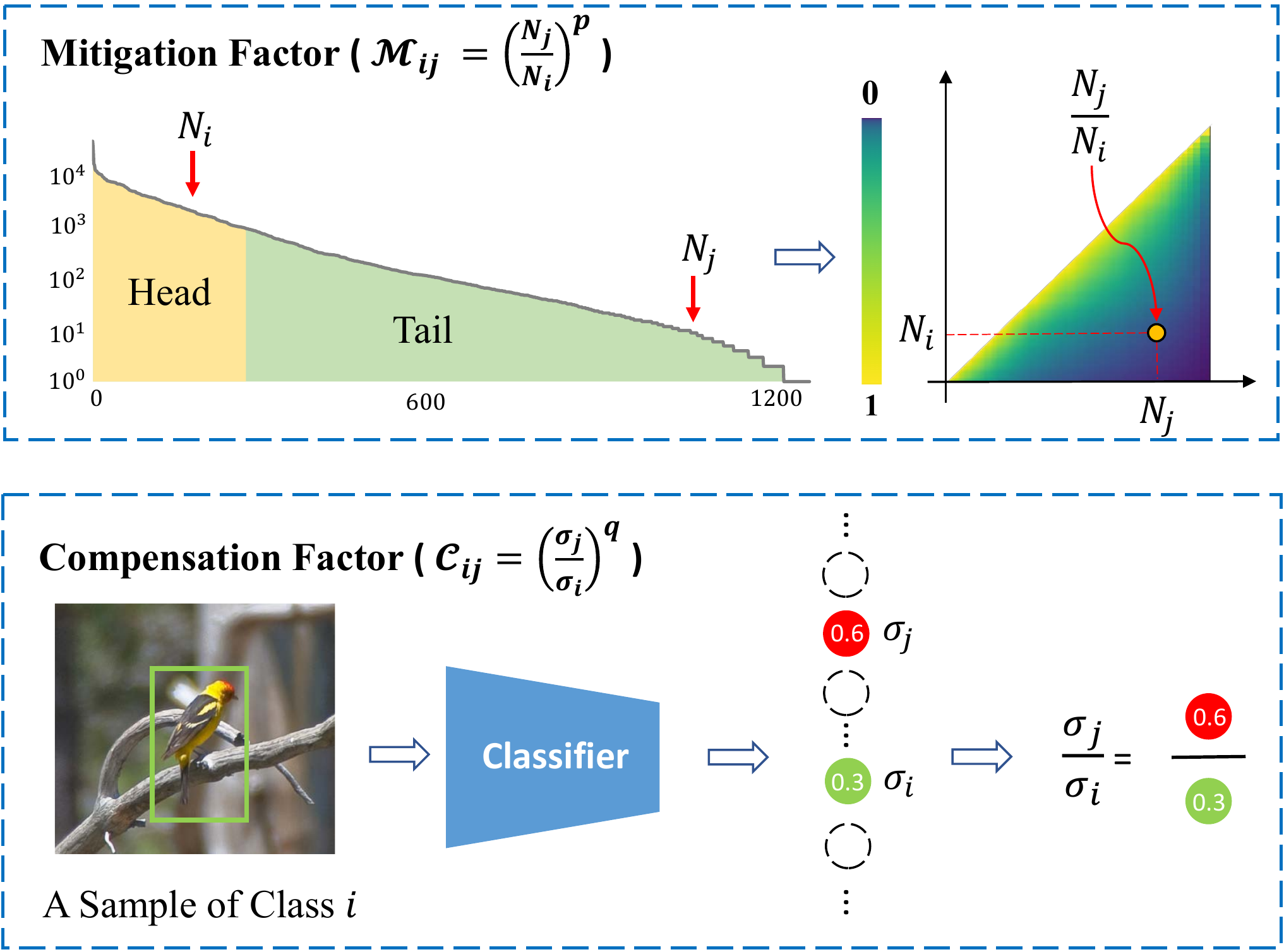}
	\end{center}
	\vspace{-10pt}
	\caption{
		\small{Seesaw Loss adjusts the punishments on tail classes with the mitigation factor $\cM_{i j}$ and the compensation factor $\cC_{ij}$.
			The mitigation factor decreases the punishments \wrt the ratio of instance numbers between different categories.
			The compensation factor increases the penalty of misclassified instances \wrt the ratio of classification probabilities between
			the false positive and the ground-truth category.
		}
	}
	\label{fig:edge}
	\vspace{-10pt}
\end{figure}

\noindent
\textbf{Mitigation Factor}.
Seesaw Loss accumulates instance number $N_i$ for each category $i$ at each iteration in the whole training process.
As shown in Fig.~\ref{fig:edge}, given an instance with positive label $i$, for another category $j$,
the mitigation factor adjusts the penalty for negative label $j$ \wrt the ratio $\frac{N_j}{N_i}$
\begin{equation} \label{eql:class-wise}
	\begin{aligned}
		\cM_{i j} =\left\{\begin{array}{ll}
			\quad 1,                        & \text { if } N_i \leq N_j \\
			\left(\frac{N_j}{N_i}\right)^p, & \text { if } N_i > N_j
		\end{array}\right.
	\end{aligned}
\end{equation}
When category $i$ is more frequent than category $j$, Seesaw Loss will reduce the penalty on category $j$, which is imposed by samples of category $i$, by a factor of $\left(\frac{N_j}{N_i}\right)^p$.
Otherwise, Seesaw Loss will keep the penalty on negative classes to reduce misclassification.
The exponent $p$ is a hyper-parameter that adapts the magnitude of mitigation.

Note that Seesaw Loss accumulates the instance numbers during training, rather than get the statistics from the whole dataset ahead of time.
This strategy brings two benefits.
First, it can be applied when the distribution of the whole training set is unavailable, \eg, training examples are obtained from a stream.
Second, the training samples of each category can be affected by the adopted data sampler~\cite{gupta2019lvis}, and the online accumulation is robust to sampling methods.
During training, the mitigation factor is uniformly initialized and smoothly updated to approximate the real data distribution.

\noindent
\textbf{Compensation Factor}.
The mitigation factor effectively balances the gradients of head and tail classes. Nevertheless,
it may cause more false positives for tail classes due to less penalty.
Moreover, the false positives cannot be eliminated by simply adjusting $p$ in $\cM_{ij}$, since it is applied to the whole category.
We propose a compensation factor that focuses on misclassified samples instead of adjusting the whole category.
As shown in Fig.~\ref{fig:edge}, this factor compensates the diminished gradient when there is misclassification, \ie, the predicted probability $\sigma_j$ of negative label $j$ is greater than $\sigma_i$.
The compensation factor $\cC_{ij}$ is calculated as
\begin{equation} \label{eql:ins-wise}
	\begin{aligned}
		\cC_{i j} =\left\{\begin{array}{ll}
			\quad 1,                                  & \text { if } \sigma_j \leq \sigma_i \\
			\left(\frac{\sigma_j}{\sigma_i}\right)^q, & \text { if } \sigma_j > \sigma_i
		\end{array}\right.
	\end{aligned}
\end{equation}
For a training sample with positive label $i$, if the predicted probability of any negative class $j$ is greater than class $i$, \ie, $\sigma_j > \sigma_i$,
the compensation factor increases the punishment on class $j$ by a factor of $\left(\frac{\sigma_j}{\sigma_i}\right)^q$,
where $q$ is a hyper-parameter to control the scale. Otherwise, $\cC_{ij}=1$ and only the mitigation factor $\cM_{i j}$ is applied.

\noindent
\textbf{Normalized Linear Activation}. The classifier in an object detector~\cite{ren2015faster} usually
predicts classification logits as $z = \cW^Tx + b$ on the dataset with balanced category distribution~\cite{lin2014coco},
where $\cW$ and $b$ are the weights and bias of the linear layer and $x$ is the input features.
On long-tailed datasets, previous works~\cite{Kang2020Decoupling, groupsoftmax} find that the weight norm of $\cW_i$ is highly related to the number of training instances in the corresponding category $i$. The
more training samples of category $i$ there are, the larger $\left \| \cW_i \right \|$ will be. This phenomenon is also observed in the feature norm $\left \| x \right \|$.
Therefore, we adopt a normalized linear activation (which~\cite{gidaris2018dynamic,OLTR,Wang_2018_cosface} are related) to re-balance the scale of $\left \| \cW_i \right \|$ and $\left \| x \right \|$ as
$z = \tau{\widetilde{\cW}}^T \widetilde{x} + b$,
where $\widetilde{\cW}_{i}=\frac{\cW_{i}}{\left\|\cW_{i}\right\|_2}, i \in C$, $\widetilde{x} = \frac{x}{\left\|x\right\|_2}$,
and $\tau$ is a temperature factor.
The normalized linear activation normalizes the weights $\cW$ and features $x$ by their $L_2$ norm to reduce their scale variance for different categories.
Thus, it effectively balances the distribution of predicted probabilities of different categories and improves the performance on a long-tailed dataset.

\subsection{Model Design for Instance Segmentation}\label{sec:modeldesign}

\noindent\textbf{Objectness Branch.} \label{sec:obj}
In contrast to image classification,
the classifier in an object detector has two functionalities.
It first determines if a bounding box is a foreground object then distinguishes which category the foreground instance belongs to.
Previous practices~\cite{cai2019cascadercnn, He_2017, ren2015faster} usually regard the background as an auxiliary category in the classifier.
Given a dataset with $C$ categories, the classifier in most detectors~\cite{He_2017,ren2015faster} predicts logits of $C+1$ classes.
Although widely adopted, this design brings difficulty when adopting Seesaw Loss to balance long-tailed distribution.
In general, most object candidates in a detector are backgrounds.
Thus, all foreground categories are much rarer categories compared to the background category.
Consequently, Seesaw Loss will significantly reduce punishments on all foreground categories.
As a result, the classifier tends to misclassify more backgrounds as foregrounds and harms the performance.

To tackle this problem, we decouple the two functionalities of the classifier in an object detecter.
Specifically, apart from the classifier with $C$ classes, we adopt an extra objectness branch to distinguish the foregrounds and backgrounds.
The objectness branch adopts the normalized linear activation to predict logits of two classes, \ie, foreground and background, and is trained by cross-entropy loss.
During inference, both the classification logit $z^{class}_i$ of category $i\in C$ and logit of objectness $z^{obj}$ are activated with a softmax function.
The final detection probability $\sigma^{det}_{i}$ for a bounding box of category $i$ is
$\sigma^{det}_i = \sigma^{class}_i \cdot \sigma^{obj}$.

\noindent\textbf{Normalized Mask Predication.} \label{sec:mask}
Inspired by normalized linear activation, we further present a normalized mask prediction to alleviate the biased training process in mask head.
In Mask R-CNN~\cite{He_2017}, a 1x1 convolution layer is applied in the end of the mask head, and the predicted logits are activated by a sigmoid function.
We normalize the weights $\cW$ of the 1x1 convolution layer and the input features $\cX$ with $L2$ normalization.
Note that the spatial size of $\cX$ is $H \times W$, we denote the feature at $(y, x)$ as $\cX_{y,x}$.
The formula of normalized mask prediction is
$z = \tau\widetilde{\cW} \ast \widetilde{\cX} + b$,
where $\widetilde{\cW}_{i}=\frac{\cW_{i}}{\left\|\cW_{i}\right\|_2}, i \in C$,
$\widetilde{\cX}_{y,x} = \frac{\cX_{y,x}}{\left\|\cX_{y,x}\right\|_2}, y \in H, x \in W$ and $\tau$ is a temperature factor.

% !TEX root = ../main.tex
\begin{table*}[t]
	\caption{
		\small{Performance comparison of Mask R-CNN~\cite{He_2017} and Cascade Mask R-CNN~\cite{cai2019cascadercnn} with Cross-Entropy (CE) Loss, Equalization Loss (EQL)~\cite{EQL}, Balanced Group Softmax (BAGS)~\cite{groupsoftmax}, and Seesaw Loss on LVIS v1 dataset~\cite{gupta2019lvis}.
			The ResNet-101~\cite{He_2016} w/ FPN~\cite{lin2017_fpn} is adopted as backbone. All models are trained with random sampler or repeat factor sampler (RFS)~\cite{gupta2019lvis} by 2x schedule in an end-to-end pipeline.
			Norm Mask indicates the proposed Normalized Mask Prediction in Sec.~\ref{sec:mask}.
		}
	}\label{tab:results}
	\vspace{-20pt}
	\begin{center}
		\addtolength\tabcolsep{-0.1em}
		\scalebox{0.95}{
			\begin{tabular}{c|c|c|c|c|ccc|c}
				\hline
				Framework                                                     & Sampler                                   & Loss                                               & Split                     & $AP$          & $AP_{r}$ & $AP_{c}$ & $AP_{f}$ & $AP^{box}$ \\ \hline
				\multirow{5}{*}{Mask R-CNN~\cite{He_2017}}                    & \multirow{5}{*}{Random}                   & Cross-Entropy (CE)                                 & \multirow{5}{*}{val}      & 20.6          & 0.8      & 19.3     & 30.7     & 21.7       \\
				                                                              &                                           & Equalization Loss (EQL)~\cite{EQL}                 &                           & 22.7          & 3.7      & 23.3     & 30.4     & 24.0       \\
				                                                              &                                           & Balanced Group Softmax (BAGS)~\cite{groupsoftmax}  &                           & 25.6          & 17.3     & 25.0     & 30.1     & 26.4       \\
				                                                              &                                           & Seesaw Loss                                        &                           & 26.6          & 18.1     & 25.8     & 31.2     & 27.4       \\
				                                                              &                                           & Seesaw Loss + Norm Mask                            &                           & \textbf{27.1} & 18.7     & 26.3     & 31.7     & 27.4       \\ \hline
				\multirow{5}{*}{Mask R-CNN~\cite{He_2017}}                    & \multirow{5}{*}{RFS~\cite{gupta2019lvis}} & Cross-Entropy (CE)                                 & \multirow{5}{*}{val}      & 25.5          & 16.6     & 24.5     & 30.6     & 26.6       \\
				                                                              &                                           & Equalization Loss (EQL)~\cite{EQL}                 &                           & 26.2          & 17.0     & 26.2     & 30.2     & 27.6       \\
				                                                              &                                           & Balanced Group Softmax (BAGS)~\cite{groupsoftmax}  &                           & 25.8          & 16.5     & 25.7     & 30.1     & 26.5       \\
				                                                              &                                           & Seesaw Loss (Ours)                                 &                           & 27.6          & 20.6     & 27.3     & 31.1     & 28.9       \\
				                                                              &                                           & Seesaw Loss + Norm Mask (Ours)                     &                           & \textbf{28.1} & 20.0     & 28.0     & 31.8     & 28.9       \\
				\hline
				\hline
				\multirow{5}{*}{Cascade Mask R-CNN~\cite{cai2019cascadercnn}} & \multirow{5}{*}{Random}                   & Cross-Entropy (CE)                                 & \multirow{5}{*}{val}      & 22.6          & 2.4      & 22       & 32.2     & 25.5       \\
				                                                              &                                           & Equalization Loss (EQL)~\cite{EQL}                 &                           & 24.3          & 5.1      & 25.3     & 31.7     & 27.3       \\
				                                                              &                                           & Balanced Group Softmax (BAGS)~\cite{groupsoftmax}  &                           & 27.9          & 19.6     & 27.7     & 31.6     & 31.5       \\
				                                                              &                                           & Seesaw Loss (Ours)                                 &                           & 29.0          & 21.1     & 28.6     & 33.0     & 32.8       \\
				                                                              &                                           & Seesaw Loss + Norm Mask (Ours)                     &                           & \textbf{29.6} & 20.3     & 29.3     & 34.0     & 32.7       \\ \hline
				\multirow{5}{*}{Cascade Mask R-CNN~\cite{cai2019cascadercnn}} & \multirow{5}{*}{RFS~\cite{gupta2019lvis}} & Cross-Entropy (CE)                                 & \multirow{5}{*}{val}      & 27.0          & 16.6     & 26.7     & 32.0     & 30.3       \\
				                                                              &                                           & Equalization Loss (EQL)~\cite{EQL}                 &                           & 27.1          & 17.0     & 27.2     & 31.4     & 30.4       \\
				                                                              &                                           & Balanced Group Softmax  (BAGS)~\cite{groupsoftmax} &                           & 27.0          & 16.9     & 26.9     & 31.7     & 30.2       \\
				                                                              &                                           & Seesaw Loss  (Ours)                                &                           & 29.3          & 21.7     & 29.2     & 32.8     & 32.8       \\
				                                                              &                                           & Seesaw Loss + Norm Mask (Ours)                     &                           & \textbf{30.1} & 21.4     & 30.0     & 33.9     & 32.8       \\
				\hline
				\hline
				\multirow{2}{*}{Mask R-CNN~\cite{He_2017}}                    & \multirow{2}{*}{RFS~\cite{gupta2019lvis}} & Cross-Entropy (CE)                                 & \multirow{2}{*}{test-dev} & 25.1          & 13.0     & 24.8     & 30.8     & -          \\
				                                                              &                                           & Seesaw Loss + Norm Mask (Ours)                     &                           & 27.9          & 20.3     & 27.1     & 32.2     & -          \\
				\hline
				\multirow{2}{*}{Cascade Mask R-CNN~\cite{cai2019cascadercnn}} & \multirow{2}{*}{RFS~\cite{gupta2019lvis}} & Cross-Entropy (CE)                                 & \multirow{2}{*}{test-dev} & 26.4          & 15.5     & 25.5     & 32.3     & -          \\
				                                                              &                                           & Seesaw Loss + Norm Mask (Ours)                     &                           & \textbf{30.0} & 23.0     & 29.3     & 34.1     & -          \\
				\hline
			\end{tabular}}
	\end{center}
	\vspace{-23pt}
\end{table*}

\section{Experiments}

\subsection{Experimental Settings}\label{sec:exp_set}

\noindent
\textbf{Datasets.}
We perform experiments on the challenging LVIS v1 dataset~\cite{gupta2019lvis}.
LVIS is a large vocabulary instance segmentation dataset containing 1203 categories with high-quality instance mask annotations.
LVIS v1 provides a \emph{train} split with 100k images, a \emph{val} split with 19.8k images and a \emph{test-dev} split with 19.8k images.
According to the numbers of images that each category appears in the \emph{train} split, the categories are divided into three groups: rare (1-10 images),
common (11-100 images) and frequent ($>$100 images).

\noindent
\textbf{Evaluation metrics.}
The results of instance segmentation are evaluated with $AP$ of mask prediction,
which is averaged at different IoU thresholds (from 0.5 to 0.95) across categories. The $\text{AP}$ for rare, common and frequent categories
are denoted as $\text{AP}_{r}$, $\text{AP}_{c}$ and $\text{AP}_{f}$.
The $\text{AP}$ for detection boxes is denoted as $\text{AP}^{box}$.

\noindent
\textbf{Implementation Details.}
We implement our method with mmdetection~\cite{mmdetection} and train Mask R-CNN~\cite{He_2017}, Cascade Mask R-CNN~\cite{cai2019cascadercnn}
using the 2x training schedule~\cite{mmdetection, Detectron2018}.
The model is trained with batch size of 16 for 24 epochs.
The learning rate is 0.02, and it will decrease by 0.1 after 16 and 22 epochs, respectively.
ResNet-50~\cite{He_2016} with FPN~\cite{lin2017_fpn} backbone is adopted if not further specified.
Following the practice in mmdetection~\cite{mmdetection}, we adopt multi-scale with horizontally flip augmentation during training.
Specifically, we randomly resize the shorter edge of the image within $\{$640, 672, 704, 736, 768, 800$\}$ pixels and
keep the longer edge smaller than 1333 pixels without changing the aspect ratio.
In inference, we adopt single-scale testing with image size of $1333 \times 800$ pixels and score thresholds of $10^{-3}$ without bells and whistles.

Apart from the standard random sampler that samples images in \emph{train} split randomly,
the repeat factor sampler (RFS)~\cite{gupta2019lvis, Mahajan_explore} is also evaluated in experiments.
RFS oversamples categories that appear in less than 0.1\% of the total images and is effective to improve the overall $\text{AP}$.
The ablation study is conducted with RFS if not further specified.
We adopt Seesaw Loss in the box classification branch of Mask R-CNN~\cite{He_2017} with hyper-parameter $p = 0.8$, $q = 2$, and $\tau = 20$.
In Cascade Mask R-CNN~\cite{cai2019cascadercnn}, Seesaw Loss is adopted in box classification branches of all three stages with the same hyper-parameters as that in Mask R-CNN.
We further evaluate the proposed Normalized Mask Prediction and integrate it into the mask head of Mask R-CNN and all the mask heads in Cascade Mask R-CNN.
For simplicity, Normalized Mask Prediction adopts the same temperature, \ie $\tau = 20$.
We use the \emph{train} split for training and report the performance
on \emph{val} split for ablation study.
The performance of our method is also reported on \emph{test-dev} split.

\subsection{Benchmark Results}\label{sec:ins_results}
To show the effectiveness of Seesaw Loss, we perform extensive experiments with different data samplers and
instance segmentation frameworks. We adopt Mask R-CNN~\cite{He_2017} with ResNet-101~\cite{He_2016} backbone with FPN~\cite{lin2017_fpn} and train the models
with the random sampler or the repeat factor sampler (RFS) by 2x schedule.

As shown in Table~\ref{tab:results}, Seesaw Loss significantly outperforms Cross-Entropy (CE) loss by 6.0\% $\text{AP}$ with random sampler and 2.1\% $\text{AP}$ on the
stronger baseline with RFS.
The improvements on
$\text{AP}_{r}$, $\text{AP}_{c}$, and $\text{AP}_{f}$ with both samplers reveals the effectiveness of Seesaw Loss on categories with different frequency.
We further integrate the proposed Normalized Mask Prediction (Norm Mask) into Mask R-CNN~\cite{He_2017} with Seesaw Loss.
Without extra cost, the overall $\text{AP}$ is improved from 26.6\% to 27.1 \%
and 27.6\% to 28.1\% with random sampler and RFS, respectively.

Apart from the CE loss baseline, we further compare
Seesaw Loss with recent designs for long-tailed instance segmentation, \ie, Equalization Loss (EQL)~\cite{EQL}
and Balanced Group Softmax (BAGS)~\cite{groupsoftmax}, in Table~\ref{tab:results}.
Seesaw Loss outperforms EQL by 3.9\% $\text{AP}$ and 1.4\% $\text{AP}$, and
outperforms BAGS by 1.0\% $\text{AP}$ and 1.8\% $\text{AP}$ with random sampler and RFS, respectively.
Seesaw Loss also achieves higher $\text{AP}_{r}$, $\text{AP}_{c}$ and $\text{AP}_{f}$ than the two methods consistently.
Notably, EQL and BAGS achieve lower $\text{AP}_{f}$ than the CE baseline while Seesaw Loss does not.
This phenomenon indicates that these two methods improve the performance of rare and common categories while sacrificing frequent categories.

We further compare Seesaw Loss with previous methods~\cite{groupsoftmax, EQL} with both random sampler and RFS on Cascade Mask R-CNN~\cite{cai2019cascadercnn}.
It's a representative framework of cascade methods~\cite{chen2019hybrid, cai2019cascadercnn} that outperforms Mask R-CNN~\cite{He_2017}.
As shown in Table~\ref{tab:results}, Seesaw Loss performs much superior to previous works~\cite{groupsoftmax, EQL} on Cascade Mask R-CNN.
Specifically, Seesaw Loss improves the baseline by \textbf{6.4\%} $\text{AP}$ and \textbf{2.3\%} $\text{AP}$ with random sampler and RFS, respectively.
With Normalized Mask Prediction, Cascade Mask-RCNN with Seesaw Loss finally achieves \textbf{29.6\%} $\text{AP}$ and \textbf{30.1\%} $\text{AP}$ with the two
samplers, respectively.
Moreover, Seesaw Loss is also evaluated on \emph{test-dev} split and consistently obtains significant gains over the CE baseline.

\subsection{Ablation study}
We conduct a comprehensive ablation study to verify the effectiveness of each design choice in the proposed method.

\begin{table}[t]
	\caption{
		\small{Ablation study of each design in Seesaw Loss with Mask R-CNN w/ R-50 FPN backbone and repeat
			factor sampler. MF, CF, NLA indicate
			mitigation factor, compensation factor, and normalized linear activation, respectively.
		}
	}\label{tab:seesaw_design}
	\vspace{-15pt}
	\begin{center}
		\addtolength\tabcolsep{-0.3em}
		\scalebox{1.0}{
			\begin{tabular}{ccc|c|ccc|c}
				\hline
				MF         & CF         & NLA        & $AP$ & $AP_{r}$ & $AP_{c}$ & $AP_{f}$ & $AP^{box}$ \\ \hline
				           &            &            & 23.7 & 13.5     & 22.8     & 29.3     & 24.7       \\
				\checkmark &            &            & 25.1 & 16.7     & 24.5     & 29.4     & 26.2       \\
				           & \checkmark &            & 24.1 & 13.2     & 23.5     & 29.5     & 25.1       \\
				\checkmark & \checkmark &            & 25.7 & 19.1     & 25.0     & 29.4     & 26.8       \\
				           &            & \checkmark & 24.7 & 15.0     & 24.1     & 29.6     & 25.6       \\
				\checkmark & \checkmark & \checkmark & 26.4 & 19.6     & 26.1     & 29.8     & 27.4       \\
				\hline
			\end{tabular}}
	\end{center}
	\vspace{-25pt}
\end{table}

\noindent
\textbf{Components in Seesaw Loss}. There are three components in Seesaw Loss: mitigation factor, compensation factor, and normalized linear activation.
We evaluated each component on Mask R-CNN with RFS (Table~\ref{tab:seesaw_design}).
The mitigation factor that mitigates the overwhelming punishments on rare classes leads to a significant improvement from 23.7\% $\text{AP}$
to 25.1\% $\text{AP}$. Notably, it improves the $\text{AP}_{r}$ of rare classes by 2.8\% AP.
The compensation factor increases the punishments of a class when it observes false positives on that class to reduce misclassification.
It improves the baseline by 0.4\% $\text{AP}$.
The combination of the mitigation and the compensation factors achieves 25.7\% $\text{AP}$,
outperforming the performance of mitigation factor by 0.6\% $\text{AP}$.
It reveals the effectiveness of instance-wise compensation to avoid misclassification.
The normalized linear activation is another important component in Seesaw Loss, which reduces the scale invariance of weights and features across different
categories. It improves the baseline performance from 23.7\% to 24.7\% $\text{AP}$.
Seesaw Loss combining all these three components achieves 26.4\% $\text{AP}$.

\begin{table}[t]
	\caption{
		\small{The effectiveness of the normalized linear activation in different methods. EQL and BAGS indicates
			Equalization Loss and Balanced Group Softmax Loss, respectively.}
	}\label{tab:nla}
	\vspace{-15pt}
	\begin{center}
		\addtolength\tabcolsep{-0.3em}
		\scalebox{1.0}{
			\begin{tabular}{c|c|c|ccc|c}
				\hline
				Method & NLA        & $AP$ & $AP_{r}$ & $AP_{c}$ & $AP_{f}$ & $AP^{box}$ \\ \hline
				EQL    &            & 25.1 & 17.4     & 24.8     & 28.8     & 26.1       \\
				EQL    & \checkmark & 25.4 & 17.8     & 25.2     & 29.1     & 26.5       \\
				BAGS   &            & 24.7 & 15.6     & 24.4     & 28.9     & 25.2       \\
				BAGS   & \checkmark & 25.5 & 19.2     & 25.0     & 28.9     & 25.8       \\
				Seesaw &            & 25.7 & 19.1     & 25.0     & 29.4     & 26.8       \\
				Seesaw & \checkmark & 26.4 & 19.6     & 26.1     & 29.8     & 27.4       \\
				\hline
			\end{tabular}}
	\end{center}
	\vspace{-25pt}
\end{table}
\noindent\textbf{Normalized linear activation}.
We empirically find that normalized linear activation (NLA) helps to improve the performance of both CE Loss and Seesaw Loss.
Therefore, we further integrate NLA with equalization loss (EQL) and balanced group softmax (BAGS) for fair comparisons.
Results in Table~\ref{tab:nla} show that NLA improves the performance of EQL and BAGS by 0.3\% and 0.8\% $\text{AP}$, respectively.
It is noteworthy that Seesaw Loss outperforms EQL and BAGS no matter whether NLA is adopted.

\noindent
\textbf{Cumulative Sample Numbers.}
Different from previous works~\cite{cao_2019_LDAM, EQL, groupsoftmax} that rely on the pre-computed frequency distribution of categories in the dataset,
Seesaw Loss accumulates the sample numbers of each category during training.
We compare different approaches to obtain the sample numbers of categories for Seesaw Loss (Table~\ref{tab:nums}).
Directly using the statistics of \emph{train} split decreases the performance of Seesaw Loss by 0.3\% $\text{AP}$.
The reason lies in that the data sampler, \eg, repeat factor sampler, changes the frequency distribution of categories during training.
We also explore loading the pre-recorded distribution of training samples from a model trained with Seesaw Loss.
It achieves a similar performance with accumulating the training samples online (26.3\% $\text{AP}$ \vs 26.4\% $\text{AP}$).
These results verify the effectiveness and simplicity of online accumulating.

\begin{table}[t]
	\caption{
		\small{Comparison of different approaches to obtain sample numbers of different categories in Seesaw Loss.
			From dataset indicates to obtain the distribution of instance numbers directly from the \emph{train} split.
			Pre-Record indicates to load the cumulative sample numbers from a model trained with Seesaw Loss.
			Online indicates to accumulate the sample numbers during training.
		}
	}\label{tab:nums}
	\vspace{-15pt}
	\begin{center}
		\addtolength\tabcolsep{-0.3em}
		\scalebox{1.0}{
			\begin{tabular}{c|c|ccc|c}
				\hline
				Source       & $AP$ & $AP_{r}$ & $AP_{c}$ & $AP_{f}$ & $AP^{box}$ \\ \hline
				From dataset & 26.1 & 19.7     & 25.6     & 29.5     & 27.2       \\
				Pre-Record   & 26.3 & 19.6     & 25.8     & 29.8     & 27.4       \\
				Online       & 26.4 & 19.6     & 26.1     & 29.8     & 27.4       \\
				\hline
			\end{tabular}}
	\end{center}
	\vspace{-23pt}
\end{table}

\begin{table}[t]
	\caption{
		\small{Ablation study of the hyper-parameter $p$ in $\left(\frac{N_j}{N_i}\right)^p$ of mitigation factor.
			The normalized linear activation is not adopted in this table.
			$p=0.8$ is the default setting in other experiments.}
	}\label{tab:p}
	\vspace{-15pt}
	\begin{center}
		\addtolength\tabcolsep{-0.3em}
		\scalebox{1.0}{
			\begin{tabular}{c|c|ccc|c}
				\hline
				$p$          & $AP$ & $AP_{r}$ & $AP_{c}$ & $AP_{f}$ & $AP^{box}$ \\ \hline
				0.2          & 24.4 & 14.7     & 23.6     & 29.4     & 25.4       \\
				0.4          & 24.9 & 15.2     & 24.6     & 29.6     & 26.0       \\
				0.6          & 25.4 & 17.9     & 24.6     & 29.5     & 26.5       \\
				\textbf{0.8} & 25.7 & 19.1     & 25.0     & 29.4     & 26.8       \\
				1.0          & 25.5 & 17.6     & 25.2     & 29.2     & 26.4       \\
				1.2          & 25.3 & 18.1     & 24.7     & 29.0     & 26.5       \\
				\hline
			\end{tabular}}
	\end{center}
	\vspace{-23pt}
\end{table}

\begin{table}[]
	\caption{
		\small{Ablation study of the hyper-parameter $q$ in $\left(\frac{\sigma_j}{\sigma_i}\right)^q$ of compensation factor.
			The normalized linear activation is not adopted in this table.
			$q=2.0$ is the default setting. }
	}\label{tab:q}
	\vspace{-15pt}
	\begin{center}
		\addtolength\tabcolsep{-0.3em}
		\scalebox{1.0}{
			\begin{tabular}{c|c|ccc|c}
				\hline
				$q$          & $AP$ & $AP_{r}$ & $AP_{c}$ & $AP_{f}$ & $AP^{box}$ \\ \hline
				% 0            & 25.1 & 16.7     & 24.5     & 29.4     & 26.2       \\
				0.5          & 25.4 & 17.6     & 25.0     & 29.4     & 26.3       \\
				1.0          & 25.5 & 17.5     & 25.1     & 29.4     & 26.6       \\
				1.5          & 25.5 & 17.8     & 25.1     & 29.3     & 26.8       \\
				\textbf{2.0} & 25.7 & 19.1     & 25.0     & 29.4     & 26.8       \\
				2.5          & 25.6 & 17.7     & 25.2     & 29.4     & 26.5       \\
				3.0          & 25.4 & 17.3     & 24.9     & 29.5     & 26.4       \\
				\hline
			\end{tabular}}
	\end{center}
	\vspace{-25pt}
\end{table}

\noindent
\textbf{Hyper-parameters}. We study the hyper-parameters, \ie, $p$, $q$, $\tau$, adopted in different components of Seesaw Loss.
The normalized linear activation is not applied when studying the mitigation and compensation factors (25.7\% AP with this setting).
In Table~\ref{tab:p}, we explore $p$ in $\left(\frac{N_j}{N_i}\right)^p$ of mitigation factor.
$p$ controls the magnitude to mitigate the punishments on rare classes.
A higher value of $p$ will reduce punishments more, as well as increase the risk of inducing false positives of tail classes.
Therefore, it is critical to find a suitable $p$.
Results show that $p=0.8$ achieves the best performance.
In Table~\ref{tab:q}, we explore $q$ in $\left(\frac{\sigma_j}{\sigma_i}\right)^q$ of the compensation factor.
$q$ controls the magnitude to compensate the reduced punishments on tail classes when false positives are observed.
We study the effectiveness of different $q$ and find $q=2.0$ achieves the best performance.
Notably, $q$ is robust across different values as the best value is only 0.3\% AP better than the worst value.
In Table~\ref{tab:tau}, we study the temperature $\tau$ in normalized linear activation (NLA).
$\tau$ determines the variance of the classifier's predicted logit $z$. If $\tau$ is too small, the variance of $z$ is insufficient to
distinguish positive and negative samples. However, if $\tau$ is too big, the target of balancing the variance in weights and
features between different categories will be sacrificed.
We choose $\tau=20$ in the NLA as it achieves the best performance.

\noindent
\textbf{Objectness Branch.} In a common practice of object detection, the classifier predicts $C+1$ scores for $C$ foregrounds categories and one background category.
Due to the extremely imbalanced distribution between foregrounds and backgrounds,
Seesaw Loss will tend to misclassify more backgrounds as foregrounds with this design.
Thus, we adopt an extra objectness branch as described in Sec.~\ref{sec:obj}.
The results in Table~\ref{tab:obj} shows that the objectness branch does not improve the Coss-Entropy loss baseline
but is critical to Seesaw Loss.
The objectness branch helps to avoid reducing the backgrounds' punishments on $C$ foreground categories.
As a result, the objectness branch brings gains on Seesaw Loss across categories with different frequency,
and improves the overall $\text{AP}$ from 25.3\% to 26.4\%.

\begin{table}[t]
	\caption{
		\small{Ablation study of the temperature term $\tau$ in Normalized Linear Activation.
			$\tau=20$ is the default setting. }
	}\label{tab:tau}
	\vspace{-15pt}
	\begin{center}
		\scalebox{1.0}{
			\begin{tabular}{c|c|ccc|c}
				\hline
				$\tau$      & $AP$ & $AP_{r}$ & $AP_{c}$ & $AP_{f}$ & $AP^{box}$ \\ \hline
				10          & 24.7 & 16.7     & 24.1     & 28.8     & 25.7       \\
				15          & 26.0 & 19.0     & 25.5     & 29.6     & 27.0       \\
				\textbf{20} & 26.4 & 19.6     & 26.1     & 29.8     & 27.4       \\
				25          & 26.2 & 19.2     & 25.7     & 29.9     & 27.4       \\
				30          & 25.9 & 16.4     & 26.2     & 29.8     & 27.0       \\
				\hline
			\end{tabular}}
	\end{center}
	\vspace{-20pt}
\end{table}

\begin{table}[t]
	\caption{
		\small{Ablation study of the objectness branch in CE loss baseline and Seesaw Loss.
			OBJ indicates whether the objectness branch is adopted.
		}
	}\label{tab:obj}
	\vspace{-15pt}
	\begin{center}
		\addtolength\tabcolsep{-0.3em}
		\scalebox{1.0}{
			\begin{tabular}{c|c|c|ccc|c}
				\hline
				Method & OBJ        & $AP$ & $AP_{r}$ & $AP_{c}$ & $AP_{f}$ & $AP^{box}$ \\ \hline
				CE     &            & 24.0 & 14.0     & 23.4     & 29.0     & 24.9       \\
				CE     & \checkmark & 23.7 & 13.5     & 22.8     & 29.3     & 24.7       \\ \hline
				Seesaw &            & 25.3 & 16.0     & 25.1     & 29.4     & 26.5       \\
				Seesaw & \checkmark & 26.4 & 19.6     & 26.1     & 29.8     & 27.4       \\
				\hline
			\end{tabular}}
	\end{center}
	\vspace{-25pt}
\end{table}

\noindent
\textbf{Training Pipeline}. Apart from the end-to-end training pipeline, we further explore the popular
decoupling training pipeline~\cite{Kang2020Decoupling, groupsoftmax} on Mask R-CNN.
Specifically, we pre-train the Mask R-CNN with Cross-Entropy loss using either random
sampler or repeat factor sampler for 2x schedule.
Then we finetune the final fully-connected layer of the classifier with all other components fixed. The 1x schedule and repeat
factor sampler is adopted during finetuning.
As shown in Table~\ref{tab:fine}, Seesaw Loss with the pre-trained model on repeat factor sampler (P-RFS)
achieves 25.8\% $\text{AP}$, outperforming other methods with decoupling training pipeline.
Notably, Seesaw Loss performs better with the end-to-end training pipeline than with the decoupling training pipeline.
It indicates Seesaw Loss provides a simpler and more effective solution to long-tailed instance segmentation without relying on complex training pipelines.

\begin{table}[t]
	\caption{
		\small{Explorations of decouple training pipelines~\cite{Kang2020Decoupling, groupsoftmax} for instance segmentation. Mask R-CNN with different
			classification loss is finetuned for 1x schedule with RFS sampler.
			We adopt pre-trained models with random (P-Rand) or repeat factor sampler (P-RFS) for 2x schedule.
			`DE-' indicates the model is trained with decoupling training pipelines.
		}
	}\label{tab:fine}
	\addtolength\tabcolsep{-0.45em}
	\vspace{-15pt}
	\begin{center}
		\scalebox{1.0}{
			\begin{tabular}{c|cc|c|ccc|c}
				\hline
				Loss      & P-Rand     & P-RFS      & $AP$ & $AP_{r}$ & $AP_{c}$ & $AP_{f}$ & $AP^{box}$ \\ \hline
				EQL       &            &            & 25.1 & 17.4     & 24.8     & 28.8     & 26.1       \\
				BAGS      &            &            & 24.7 & 15.6     & 24.4     & 28.9     & 25.2       \\
				Seesaw    &            &            & 26.4 & 19.6     & 26.1     & 29.8     & 27.4       \\ \hline
				DE-EQL    & \checkmark &            & 23.9 & 12.9     & 23.7     & 28.9     & 25.4       \\
				DE-EQL    &            & \checkmark & 25.2 & 15.9     & 25.5     & 28.9     & 26.5       \\
				DE-BAGS   & \checkmark &            & 25.4 & 16.3     & 25.1     & 29.7     & 26.3       \\
				DE-BAGS   &            & \checkmark & 25.6 & 17.0     & 25.3     & 29.8     & 26.6       \\
				DE-Seesaw & \checkmark &            & 25.1 & 16.6     & 24.6     & 29.5     & 26.2       \\
				DE-Seesaw &            & \checkmark & 25.8 & 18.7     & 25.3     & 29.6     & 26.9       \\
				\hline
			\end{tabular}}
	\end{center}
	\vspace{-25pt}
\end{table}

\subsection{Long-Tailed Image Classification}\label{sec:cls_results}

To show the versatility of Seesaw Loss, we apply it for long-tailed image classification task on ImageNet-LT~\cite{OLTR} dataset.
ImageNet-LT~\cite{OLTR} is generated from the ImageNet-2012~\cite{ILSVRC15} dataset with long-tailed distributed categories in training set.
There are 115.8k images of 1000 categories with a maximum number of 1280 images and a minimum number of 5 images.
The performance is evaluated with top-1 accuracy on all categories and the accuracies for Many Shot ($>$ 100 images), Medium Shot (20$\sim$100 images) and
Few Shot ($<$ 20 images) categories are also reported.

We adopt two training pipelines: end-to-end training and decoupling training\cite{Kang2020Decoupling}.
We use ResNeXt-50~\cite{Xie_2017} backbone and SGD optimizer with momentum of 0.9, initial learning rate of 0.2, batch size of 512,
and cosine learning rate~\cite{SGDR} following~\cite{Kang2020Decoupling}.
For the end-to-end training pipeline, the model is trained for 90 epochs.
For decouple training pipeline~\cite{Kang2020Decoupling}, we load the pre-trained ResNeXt-50~\cite{Xie_2017} with Cross-Entropy Loss (CE),
and finetune the classifier with class-balanced sampler while fixing all other layers for 10 epochs.
Seesaw Loss in image classification mostly follows the hyper-parameters on the instance
segmentation task except for $q$ in the compensation factor. We adopt $q=1$ for ImageNet-LT dataset. The study of $q$ for ImageNet-LT dataset is shown in
Table~\ref{tab:cls_q}.

We report the performance of Seesaw Loss in Table~\ref{tab:cls}.
Seesaw Loss improves top-1 accuracies of CE from 44.4\% to 49.7\% and 50.4\% with the decoupling training and the end-to-end training pipeline, respectively.
Similar to our observations on the instance segmentation task, Seesaw Loss performs better with the end-to-end training pipeline on image classification.
The performance achieved by Seesaw Loss with the end-to-end pipeline is competitive among previous methods on ImageNet-LT~\cite{OLTR}.

\begin{table}[t]
	\caption{
		\small{Comparison of different methods on ImageNet-LT~\cite{OLTR} test set.
			ResNeXt-50~\cite{Xie_2017} backbone is adopted in experiments.
			Decouple means using decouple training pipeline~\cite{Kang2020Decoupling}.}
	}\label{tab:cls}
	\addtolength\tabcolsep{-0.35em}
	\vspace{-15pt}
	\begin{center}
		\scalebox{1.0}{
			\begin{tabular}{c|c|c|ccc}
				\hline
				Method                                & Decouple   & Overall & Many & Medium & Few  \\ \hline
				CE                                    &            & 44.4    & 65.9 & 37.5   & 7.7  \\
				Focal Loss~\cite{lin2017_focal}       &            & 43.3    & 64.5 & 36.3   & 7.8  \\
				CB-Focal~\cite{Cui_2019}              &            & 45.3    & 60.4 & 40.6   & 19.2 \\
				EQL~\cite{Cui_2019}                   &            & 46.0    & 61.7 & 42.5   & 13.8 \\
				NCM~\cite{Kang2020Decoupling}         & \checkmark & 47.3    & 56.6 & 45.3   & 28.1 \\
				cRT~\cite{Kang2020Decoupling}         & \checkmark & 49.6    & 61.8 & 46.2   & 27.4 \\
				$\tau$-norm~\cite{Kang2020Decoupling} & \checkmark & 49.4    & 59.1 & 46.9   & 30.7 \\
				LWS~\cite{Kang2020Decoupling}         & \checkmark & 49.9    & 60.2 & 47.2   & 30.3 \\ \hline
				Seesaw                                & \checkmark & 49.7    & 60.7 & 46.8   & 28.9 \\
				Seesaw                                &            & 50.4    & 67.1 & 45.2   & 21.4 \\
				\hline
			\end{tabular}}
	\end{center}
	\vspace{-20pt}
\end{table}

\begin{table}[t]
	\caption{
		\small{Comparison of hyper-parameter $q$ in the compensation factor $\left(\frac{\sigma_j}{\sigma_i}\right)^q$ on ImageNet-LT~\cite{OLTR}.
		}
	}\label{tab:cls_q}
	\addtolength\tabcolsep{-0.25em}
	\vspace{-15pt}
	\begin{center}
		\scalebox{1.0}{
			\begin{tabular}{c|c|ccc}
				\hline
				$q$          & Overal & Many & Medium & Few  \\ \hline
				0.5          & 49.6   & 66.2 & 44.4   & 20.9 \\
				\textbf{1.0} & 50.4   & 67.1 & 45.2   & 21.4 \\
				1.5          & 49.6   & 66.4 & 44.3   & 20.7 \\
				2.0          & 49.4   & 66.5 & 44.0   & 20.3 \\
				2.5          & 48.4   & 65.8 & 42.9   & 18.7 \\
				\hline
			\end{tabular}}
	\end{center}
	\vspace{-25pt}
\end{table}

\vspace{-5pt}
\section{Conclusion}
In this paper, we propose Seesaw Loss for long-tailed instance segmentation.
Seesaw Loss dynamically re-balances gradients of positive and negative samples
for each category with two complementary factors.
The mitigation factor reduces punishments to tail categories \wrt the ratio of cumulative training instances between categories.
Meanwhile, the compensation factor increases the penalty of misclassified instances to avoid false positives.
Experimental results demonstrate that Seesaw Loss
provides a simpler and more effective solution to long-tailed instance segmentation without relying on complex training pipelines.

\noindent\textbf{Acknowledgements.}
This research was conducted in collaboration with SenseTime. This work is supported by GRF 14203518, ITS/431/18FX,
CUHK Agreement TS1712093, NTU NAP, A*STAR through the Industry Alignment Fund - Industry Collaboration Projects Grant,
Shanghai Committee of Science and Technology, China (Grant No. 20DZ1100800).

% !TEX root = ../main.tex
\appendix
\begin{appendices}
    \section{Analysis of $\cS_{ij}$ in Seesaw Loss}
    In this work, we propose Seesaw Loss to dynamically re-balance gradients of positive and negative samples for each category.
    Specifically, Seesaw Loss mitigates the overwhelming gradients of negative samples imposed by a head class $i$ on a tail class $j$ via decreasing the value of
    $\cS_{ij}$ in the following formula,
    \begin{equation} \label{eql:seesaw_neg_grad}
        \begin{aligned}
            \frac{\partial L_{seesaw}(\vz)} {\partial z_j} = \cS_{ij} \frac{e^{z_j}} {e^{z_i}} \widehat{\sigma}_i, \quad \\
            \text{  with  } \widehat{\sigma}_{i}=\frac{e^{z_{i}}}{\sum_{j\neq i}^{C}\cS_{ij}e^{z_{j}}+e^{z_{i}}}.
        \end{aligned}
    \end{equation}
    To further analyze the effects of adjusting the value of $\cS_{ij}$, we calculate the partial derivative
    of Eqn~\ref{eql:seesaw_neg_grad} with respect to $\cS_{ij}$ as
    \begin{equation} \label{eql:partial_dev}
        \frac{\partial(\cS_{ij} \frac{e^{z_j}} {e^{z_i}} \widehat{\sigma}_i)} {\partial\cS_{ij}} = \frac{e^{z_j}(\sum_{k\neq i, j}^{C}\cS_{ik}e^{z_{k}}+e^{z_{i}})} {(\sum_{j\neq i}^{C}\cS_{ij}e^{z_{j}}+e^{z_{i}})^2} > 0.
    \end{equation}
    The value of the partial derivative in Eqn~\ref{eql:partial_dev} is always positive.
    This indicates that the gradients of negative samples imposed by class $i$ on class $j$ will be reduced as the value of $\cS_{ij}$ decreases.

    \begin{figure}[t]
        \begin{center}
            \includegraphics[width=\linewidth]{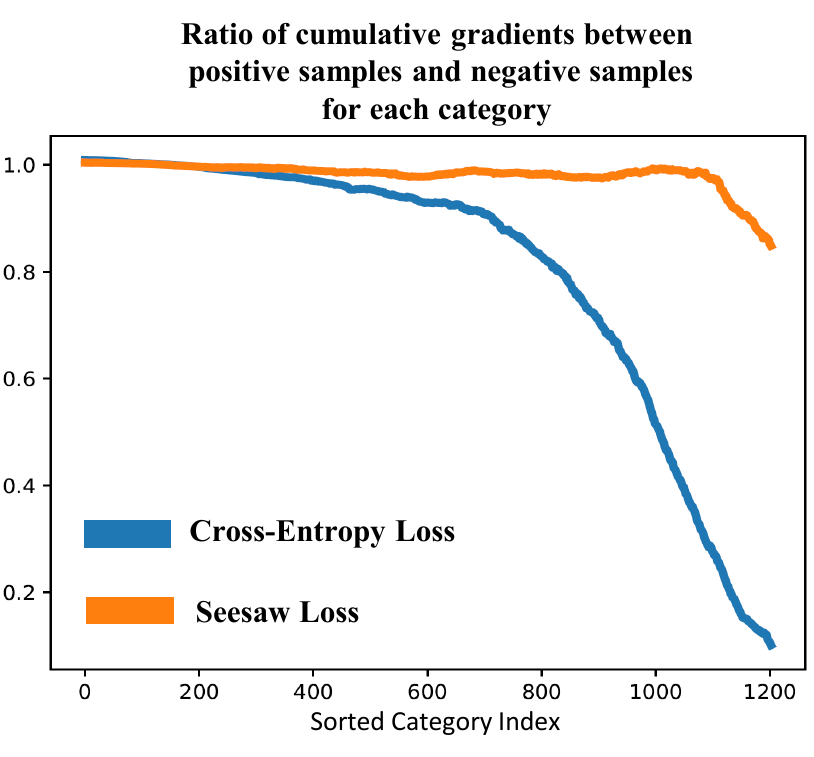}
        \end{center}
        \vspace{-15pt}
        \caption{
            \small{The distribution of the ratio of cumulative gradients between positive and negative samples for each category
                with Cross-Entropy Loss and Seesaw Loss, respectively. The categories are sorted in descending order with respect to their instance numbers.
                In contrast to Cross-Entropy Loss, Seesaw Loss effectively re-balances the gradients of
                positive and negative samples.
            }
        }
        \label{fig:grad_cum}
        \vspace{-15pt}
    \end{figure}

    \begin{figure*}[t]
        \centering
        \includegraphics[width=0.95\textwidth]{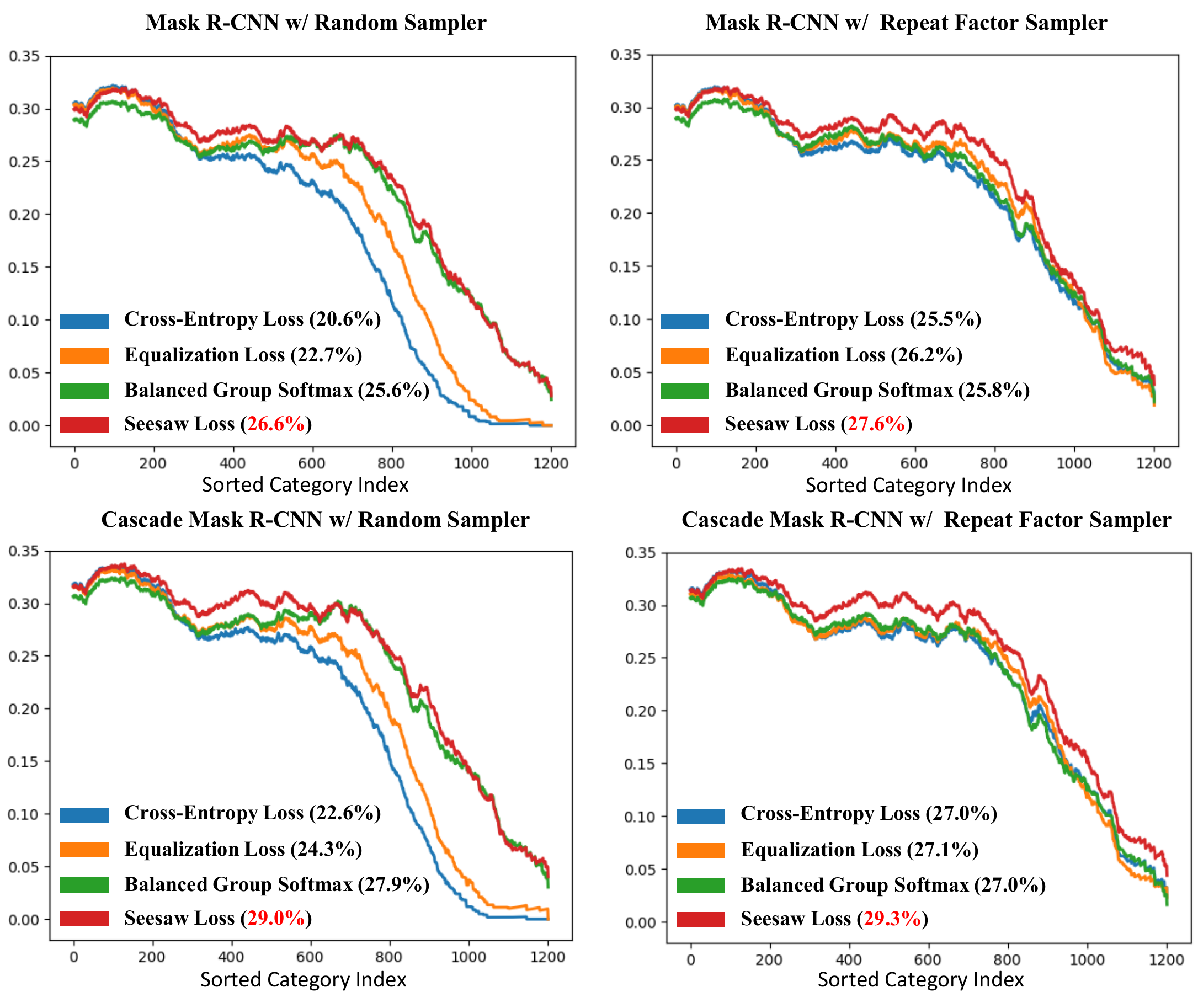}
        \vspace{-10pt}
        \caption{
            Per-category performance (AP) comparison between different methods in Table 1 of the main text. Norm Mask is not adopted for a fair comparison.
        }
        \label{fig:performance}
    \end{figure*}

    \section{How Seesaw Loss works}
    Via re-balancing gradients of positive and negative samples, Mask R-CNN~\cite{He_2017} w/ Seesaw Loss
    significantly outperforms Mask R-CNN~\cite{He_2017} w/ Cross-Entropy Loss on LVIS~\cite{gupta2019lvis} dataset.
    Here, we conduct a quantitative analysis of the effectiveness of Seesaw Loss on re-balancing the gradients of positive
    and negative samples for each category. Specifically, we adopt Mask R-CNN~\cite{He_2017} with ResNet-101~\cite{He_2016} backbone and FPN~\cite{lin2017_fpn} as
    instance segmentation framework. The Cross-Entropy Loss and Seesaw Loss are integrated into the framework and trained with random sampler by 2x schedule.
    We accumulate the gradients of positive and negative samples on predicted logit $z_i$ of each category $i$ during the whole training procedure.
    
    Figure~\ref{fig:grad_cum} shows the distribution of the ratio of cumulative gradients between positive and negative samples
    for each category in Mask R-CNN~\cite{He_2017} with Cross-Entropy Loss and Seesaw Loss, respectively.
    With Cross-Entropy Loss, tail classes obtain
    heavily imbalanced gradients of positive and negative samples during training. The overwhelming gradients of negative
    samples lead to a biased learning process for the classifier, which results in the low classification accuracy on tail classes.
    On the contrary, Seesaw Loss effectively re-balances
    the gradients of positive and negative samples across different categories. Consequently, Mask R-CNN with Seesaw Loss
    achieves significant improvements on instance segmentation performance
    as shown in Figure 1 and Table 1 in the main text.

    \section{Per-category Performance Comparison}
    In addition to the performance reported in Table 1 of the main text, we further show the per-category performance (AP) to verify the superiority of Seesaw Loss compared to other loss functions.
    As shown in Figure~\ref{fig:performance}, compared to other loss functions (\ie, Cross-Entropy Loss, Equalization Loss~\cite{EQL}, and Balanced Group Softmax~\cite{groupsoftmax}),
    Seesaw Loss consistently achieves strong performance across categories with
    different frequency on different frameworks (\ie Mask R-CNN~\cite{He_2017}, Cascade Mask R-CNN~\cite{cai2019cascadercnn}) and samplers (\ie, random sampler, repeat factor sampler~\cite{gupta2019lvis}).
    
    \begin{figure*}[t]
        \centering
        \includegraphics[width=0.95\textwidth]{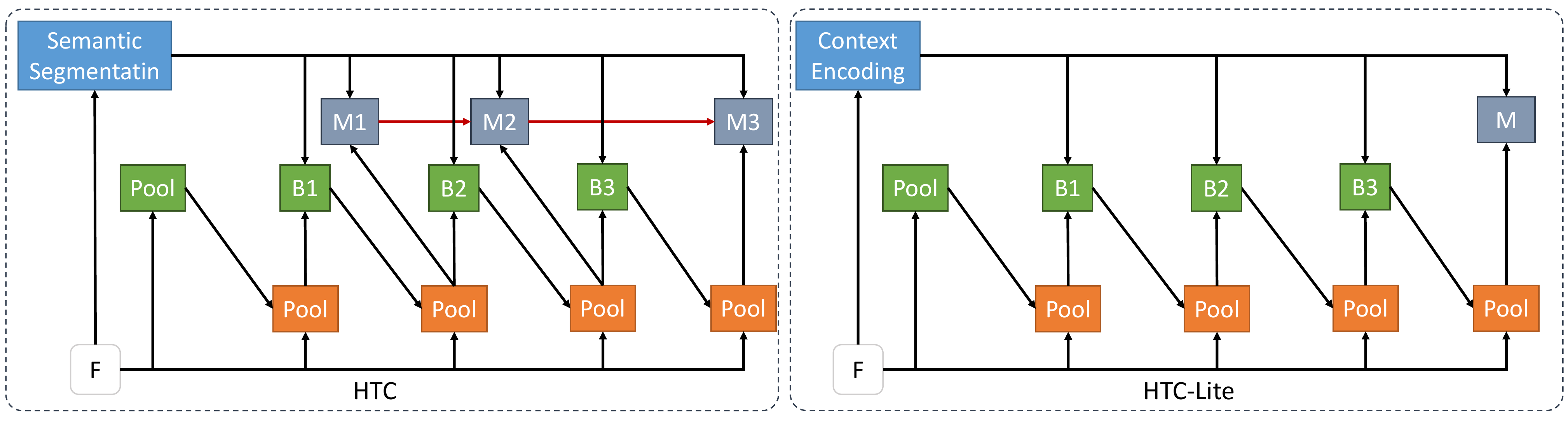}
        \vspace{-10pt}
        \caption{
            The comparison of HTC and HTC-Lite.
        }
        \label{fig:htc-lite}
    \end{figure*}

    \begin{table*}[t]
        \caption{
            \small{Step by step results of our entry on LVIS v1~\cite{gupta2019lvis} \emph{val} split.
            }
        }\label{tab:step_by_step}
        \vspace{-18pt}
        \begin{center}
            \begin{tabular}{c|c|c|ccc|c}
                \hline
                Modification                & Schedule & $AP$                                    & $AP_{r}$      & $AP_{c}$      & $AP_{f}$      & $AP^{box}$                              \\ \hline
                Mask R-CNN                  & 2x       & 18.7                                    & 1.0           & 16.1          & 29.4          & 20.1                                    \\ \hline
                + SyncBN                    & 2x       & 18.9 (+0.2)                             & 0.7           & 16.0          & 30.3          & 20.2 (+0.1)                             \\ \hline
                + CARAFE Upsample           & 2x       & 19.4 (+0.5)                             & 0.7           & 16.5          & 30.9          & 20.4 (+0.2)                             \\ \hline
                + HTC-Lite                  & 2x       & 21.9 (+2.5)                             & 1.1           & 19.8          & 33.5          & 23.6 (+3.2)                             \\ \hline
                + TSD                       & 2x       & 23.5 (+1.6)                             & 2.3           & 22.3          & 34.0          & 25.5 (+1.9)                             \\ \hline
                + Mask scoring              & 2x       & 23.9 (+0.4)                             & 2.8           & 22.4          & 35.0          & 25.6 (+0.1)                             \\ \hline
                + Training-time augmentaion & 45e      & 26.5 (+2.6)                             & 3.6           & 25.7          & 37.4          & 28.1 (+2.5)                             \\ \hline
                + Stronger neck             & 45e      & 27.0 (+0.5)                             & 3.5           & 25.8          & 38.6          & 29.1 (+1.0)                             \\ \hline
                + Stronger backbone         & 45e      & 29.9 (+2.9)                             & 4.2           & 29.4          & 41.8          & 32.1 (+3.0)                             \\ \hline
                + \textbf{Seesaw Loss}      & 45e      & 36.8 (\textbf{{\textcolor{red}{+6.9}}}) & 25.5          & 35.6          & 42.9          & 39.8 (\textbf{{\textcolor{red}{+7.7}}}) \\ \hline
                + Dual Head Classification  & 1x       & 37.3 (+0.5)                             & 26.4          & 36.3          & 43.1          & 40.6 (+0.8)                             \\ \hline
                + Test-time augmentation    & -        & \textbf{38.8} (+1.5)                    & \textbf{26.4} & \textbf{38.3} & \textbf{44.9} & \textbf{41.5} (+0.9)                    \\
                \hline
            \end{tabular}
        \end{center}
        \vspace{-24pt}
    \end{table*}

    \begin{table}[t]
        \caption{
            \small{Comparison of different cascading instance segmentation frameworks on LVIS v1~\cite{gupta2019lvis} dataset with repeat factor sampler and 1x training schedule. HTC w/o semantic indicates HTC without adopting the semantic segmentation branch
                since semantic segmentation annotations are not available on LVIS v1 dataset.}
        }\label{tab:htc-lite}
        \vspace{-15pt}
        \begin{center}
            \addtolength\tabcolsep{-0.5em}
            \scalebox{1.0}{
                \begin{tabular}{c|c|ccc|c|c}
                    \hline
                    Method                                       & $AP$ & $AP_{r}$ & $AP_{c}$ & $AP_{f}$ & $AP^{box}$ & fps \\ \hline
                    Cascade Mask R-CNN~\cite{cai2019cascadercnn} & 24.3 & 13.7     & 23.8     & 29.6     & 27.2       & 0.1 \\
                    HTC w/o semantic~\cite{chen2019hybrid}       & 24.8 & 14.5     & 24.1     & 30.2     & 27.0       & 0.1 \\
                    HTC-Lite                                     & 25.5 & 15.0     & 25.4     & 30.3     & 28.0       & 2.8 \\
                    \hline
                \end{tabular}}
        \end{center}
        \vspace{-20pt}
    \end{table}

    \section{LVIS Challenge 2020}
    
    Here we present the approach used in the entry of team \textbf{MMDet} in the LVIS Challenge 2020.
    In our entry, we adopt Seesaw Loss for long-tailed instance segmentation as described in the main text.
    Seesaw Loss improves the strong baseline by 6.9\% AP on LVIS v1 \emph{val} split.
    Furthermore, we propose HTC-Lite, a light-weight version of Hybrid Task Cascade (HTC)~\cite{chen2019hybrid} which replaces the semantic segmentation branch with a global context encoder.
    With a single model and without using external data and annotations except for standard ImageNet-1k classification dataset for backbone pre-training, our entry achieves \textbf{38.92\% AP} on the \emph{test-dev} split of the LVIS v1 benchmark.

    \subsection{HTC-Lite}\label{sec:HTC-Lite}

    We propose HTC-Lite, a light-weight version of Hybrid Task Cascade (HTC)~\cite{chen2019hybrid}, to accelerate the training and inference speed while maintaining good performance.
    As shown in Figure~\ref{fig:htc-lite}, the modifications are in two folds: replacing the semantic segmentation branch with a global context encoding branch and reducing mask heads.

    \noindent
    \textbf{Context Encoding Branch.} Since semantic segmentation annotations are not available for LVIS~\cite{gupta2019lvis} dataset,
    we replace the semantic segmentation branch with a global context encoder~\cite{Zhang_2018_CVPR} which works as a multi-label classification branch trained by a binary cross-entropy loss.
    The context encoder applies convolution layers and a global average pooling on the input feature map to obtain a feature vector.
    And an auxiliary fully connected (fc) layer is applied on the feature vector to predict the categories existing in the current image.
    By this approach, this feature vector encodes the global context information of the image. Then it is added to the RoI features used by box heads and mask heads to enrich their semantic information.

    \noindent
    \textbf{Reduced Mask Heads.}
    To further reduce the cost of instance segmentation, HTC-Lite only keeps the mask head in the last stage, which also spares the original interleaved information passing.

    In Table~\ref{tab:htc-lite}, we compare the performance and inference speed on LVIS v1~\cite{gupta2019lvis} dataset of HTC-Lite with two mainstream cascading instance segmentation frameworks, \ie, Cascade Mask R-CNN and HTC.
    The ResNet-50 with FPN backbone, repeat factor sampler and 1x training schedule are adopted in these methods. The semantic segmentation branch in HTC~\cite{chen2019hybrid} is removed
    since semantic segmentation annotations are not available on LVIS v1~\cite{gupta2019lvis} dataset. We evaluate the inference speed for each framework with a single Tesla V100 GPU. The experimental results show that HTC-Lite is not only much more efficient than its counterparts but also
    outperforms them.

    \subsection{Step by Step Results}

    We report the step-by-step results of our entry in LVIS Challenge 2020 as shown in Table~\ref{tab:step_by_step}.

    \noindent
    \textbf{Baseline.} The baseline model is Mask R-CNN~\cite{He_2017} using ResNet-50-FPN~\cite{lin2017_fpn},
    trained with multi-scale training and random data sampler by 2x schedule~\cite{mmdetection}.

    \noindent
    \textbf{SyncBN.} We use SyncBN~\cite{liu2018_panet,Peng_2018} in the backbone and heads.

    \noindent
    \textbf{CARAFE Upsample.} CARAFE~\cite{Wang_2019_ICCV} is used for upsampling in the mask head.

    \noindent
    \textbf{HTC-Lite.} We use HTC-Lite as described in Appendix~\ref{sec:HTC-Lite}.

    \noindent
    \textbf{TSD.} TSD~\cite{song2020revisiting} is used to replace the box heads in all three stages in HTC-Lite.

    \noindent
    \textbf{Mask Scoring.} We further use the mask IoU head~\cite{huang2019msrcnn} to improve mask results.

    \noindent
    \textbf{Training Time Augmentation.} We train the model with stronger augmentations with 45 epochs. The learning rate is decreased by 0.1 at 30 and 40 epochs. We randomly resize the image with its longer edge in a range of 768 to 1792 pixels. And then, we randomly crop the image to the size of $1280\times1280$ after adopting instaboost augmentation~\cite{fang2019instaboost}.

    \noindent
    \textbf{Stronger Neck.} We replace the neck architecture with an enhanced version of Feature Pyramid Grids (FPG)~\cite{chen2020feature}.
    The enhanced FPG uses deformable convolution v2 (DCNv2)~\cite{Zhu_2019_CVPR} after feature upsampling, and a downsampler version of CARAFE~\cite{Wang_2019_ICCV, wang2020carafe} for feature downsampling.

    \noindent
    \textbf{Stronger Backbone.} We use ResNeSt-200~\cite{zhang2020resnest} with DCNv2~\cite{Zhu_2019_CVPR} as the backbone.

    \noindent
    \textbf{Seesaw Loss.} We apply the proposed Seesaw Loss to classification branches of the TSD box head, in all cascading stages.
    Furthermore, we remove the original progressive constraint (PC) loss on classification branches in TSD.

    \noindent
    \textbf{Dual Head Classification.} Inspired by \cite{wang2020devil, wang2019classification}, we adopt a dual-head classification policy to further boost the performance.
    Specifically, after obtaining the model with Seesaw Loss trained by a random sampler, we freeze all components in the original model.
    Then we finetune a new classification branch for each cascading stage on the fixed model using repeat factor sampler~\cite{gupta2019lvis} by 1x schedule.
    During inference, the classification scores of original classification branches and the scores of new classification branches
    are averaged to get the final scores.

    \noindent
    \textbf{Test Time Augmentation.} We adopt multi-scale testing with horizontal flipping. Specifically, image scales are 1200, 1400, 1600, 1800, and 2000 pixels.

    \noindent
    \textbf{Final Performance on Test-dev.} After adding the abovementioned components step by step, we finally achieve \textbf{38.8\% AP} on the \emph{val} split and \textbf{38.92\% AP} on the \emph{test-dev} split.

\end{appendices}

{\small
    \bibliographystyle{ieee_fullname}
    \bibliography{sections/egbib}
}
% \clearpage

\end{document}